\documentclass[conference]{IEEEtran}
\usepackage{blindtext, graphicx}
\usepackage{color}

\ifCLASSINFOpdf
\else
\fi
\usepackage[cmex10]{amsmath}

\begin{document}
%
\title{Delta Networks for Optimized Recurrent Network Computation}

\author{\IEEEauthorblockN{Daniel Neil$^{1}$, Jun Haeng Lee$^{1,2}$, Tobi Delbruck$^{1}$, Shih-Chii Liu$^{1}$}
\IEEEauthorblockA{$^{1}$Institute of Neuroinformatics, UZH and ETH Zurich, Zurich, Switzerland\\
$^{2}$Samsung Advanced Institute of Technology, Samsung Electronics, Suwon-Si, Republic of Korea\\
Email: daniel.l.neil@ini.ethz.ch, junhaeng.lee@gmail.com, tobi@ini.ethz.ch, shih@ini.ethz.ch}
}

%


\newcommand{\dneil}[1]{\textcolor{green}{{\bf dn: }~#1}}
\newcommand{\shih}[1]{\textcolor{red}{{\bf sc: }~#1}}
\newcommand{\jhl}[1]{\textcolor{blue}{{\bf jl: }~#1}}
\newcommand{\td}[1]{\textcolor{magenta}{{\bf td:}~#1}}
\newcommand{\cg}[1]{\textcolor{cyan}{{\bf cg:}~#1}}


\maketitle

\begin{abstract}
Many neural networks exhibit stability in their activation patterns over time in response to inputs from sensors operating under real-world conditions. By capitalizing on this property of natural signals, we propose a Recurrent Neural Network (RNN) architecture called a delta network in which each neuron transmits its value only when the change in its activation exceeds a threshold. The execution of RNNs as delta networks is attractive because their states must be stored and fetched at every timestep, unlike in convolutional neural networks (CNNs). We show that a naive run-time delta network implementation offers modest improvements on the number of memory accesses and computes, but optimized training techniques confer higher accuracy at higher speedup. With these optimizations, we demonstrate a 9X reduction in cost with negligible loss of accuracy for the TIDIGITS audio digit recognition benchmark.  Similarly, on the large Wall Street Journal speech recognition benchmark even existing networks can be greatly accelerated as delta networks, and a 5.7x improvement with negligible loss of accuracy can be obtained through training.  Finally, on an end-to-end CNN trained for steering angle prediction in a driving dataset, the RNN cost can be reduced by a substantial 100X.
\end{abstract}


%
\IEEEpeerreviewmaketitle

\section{Introduction and Motivation}
\label{sec:intro}
Recurrent Neural Networks (RNNs) have achieved tremendous progress in recent years, with the increased availability of large datasets, more powerful computer resources such as GPUs, and improvements in their training algorithms.
These combined factors have enabled breakthroughs in the use of RNNs for processing of temporal sequences. Applications such as natural language processing \cite{mikolov2010recurrent}, speech recognition \cite{amodei2015deep,graves2013speech}, and attention-based models for structured prediction \cite{yao2015describing,xu2015show} have showcased the advantages of RNNs, as they provide breakthroughs in former stagnating challenges.
RNNs are attractive because they equip neural networks with memories, and the introduction of gating units such as long short-term memory (LSTM) units \cite{hochreiter1997long} and gated recurrent units (GRU) \cite{cho2014gru} has greatly improved the training process with these networks.
However, RNNs require many matrix-vector multiplications per layer to calculate the updates
of neuron activations over time.

RNNs also require a large weight memory storage that is expensive
to allocate to on-chip static random access memory.
In a 45nm technology, the energy cost of an off-chip dynamic
32-bit random access memory (SDRAM)
access is about 2nJ and the energy for a 32-bit integer multiply is about 3pJ,
so memory access is about 700 times more expensive than arithmetic \cite{horowitz2014}.
Architectures can benefit from minimizing this external memory access. Previous
work has focused on a variety of algorithmic
optimizations for reducing compute and memory access requirements for deep neural networks. These methods include reduced precision for hardware optimization (\cite{courbariaux2015binaryconnect, stromatiasneil2015robustness, courbariaux2016binarynet, esser2016convolutional, rastegari2016xnor});
weight encoding, pruning, and compression (\cite{han2015deep, han2016ese});
and architectural optimizations
(\cite{iandola2016squeezenet, szegedy2015going, huang2016deep}).
However these studies have not considered temporal properties of the data.

\begin{figure}[!t]
\centering
\includegraphics[width=0.48\textwidth]{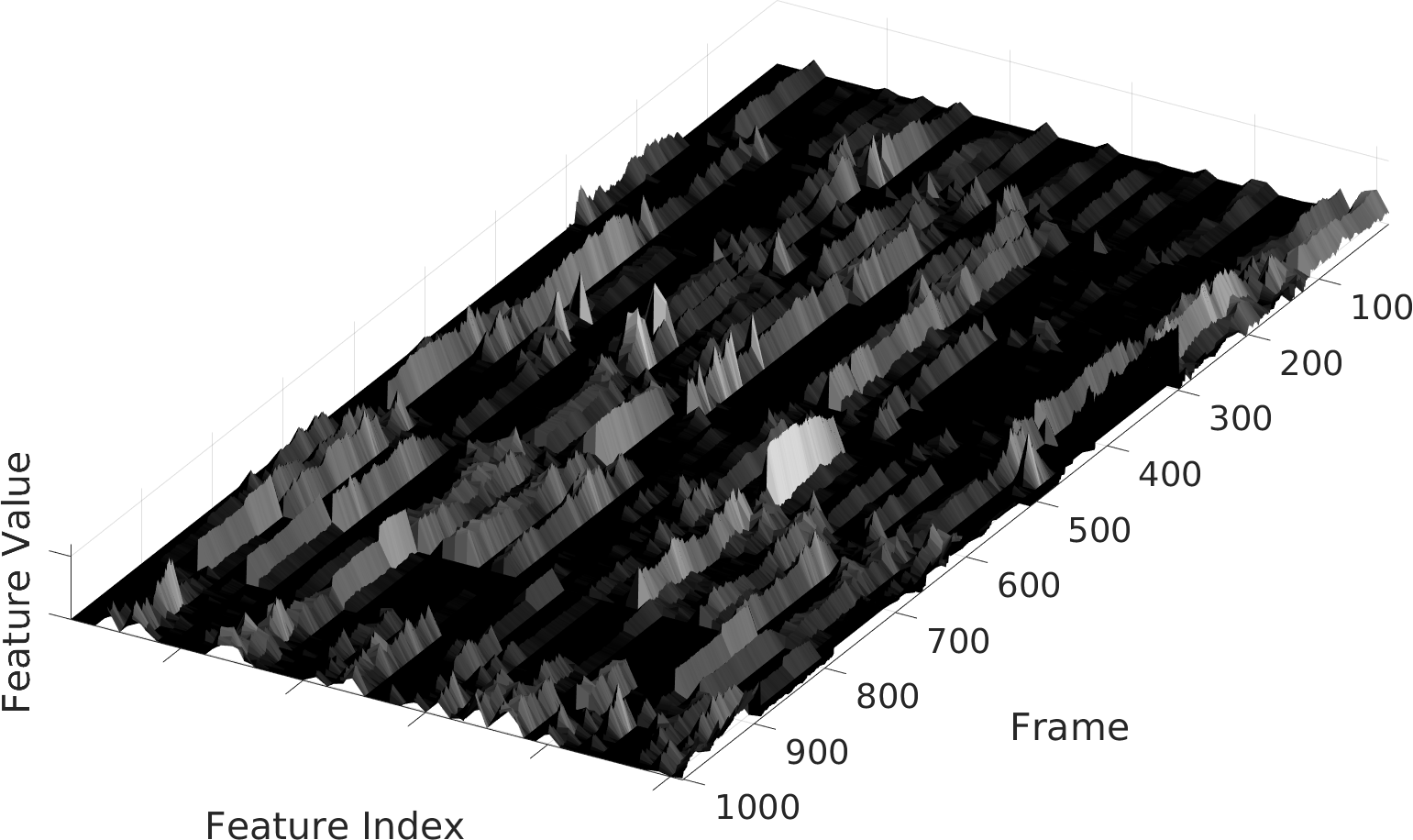}
\caption{Stability of high-level neural representations over time.  The first 1000 frames (40s) from a Hollywood-2 \cite{marszalek09} scene recognition clip (the introduction to \textit{American Beauty}) are presented to a standard convolutional network (\mbox{VGG-S)}~\cite{chatfield2014return}, with the first 50 (arbitrary) features of the top-level feature vector layer are plotted over time.  Note that peaks tend to stay relatively constant over time, showing network output consistency over time rather than random feature activation.}
\label{fig:feature_stability}
\end{figure}

\begin{figure}[!t]
\centering
\includegraphics[width=0.48\textwidth]{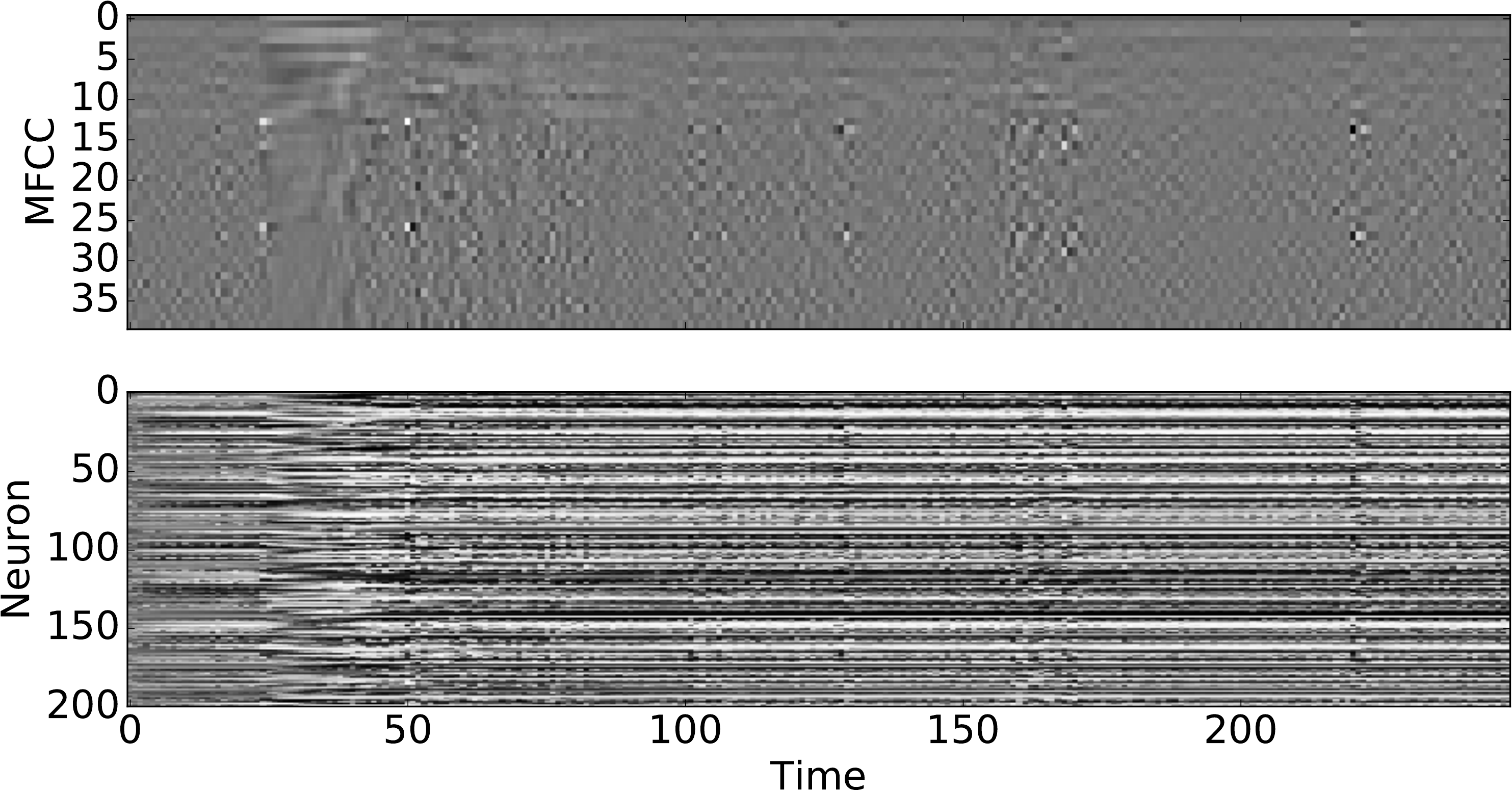}
\caption{Stability in RNN activations over time.  The top figure shows the continually-changing MFCC features for a spoken digit from the TIDIGITS dataset \cite{leonard1993tidigits}; the bottom figure shows the corresponding neural network output activations in response to these features.  Note the slow evolution of the network states over timesteps.}
\label{fig:tidigits_feature_stability}
\end{figure}

It can be observed that natural inputs to a neural network tend to have a high degree of temporal autocorrelation, resulting in slowly-changing network states.
Fig.~\ref{fig:feature_stability} demonstrates this property with a standard convolutional network (VGG-S \cite{chatfield2014return}) operating on a standard video dataset. As seen, the neural representation over time is highly redundant.  This slow changing activation feature is also seen within the computation of RNNs processing natural inputs, for example, speech (Fig.~\ref{fig:tidigits_feature_stability}).


Delta networks, as introduced here, exploit the temporal stability of both the input stream and the associated neural representation to reduce memory access and computation without loss of accuracy.
By caching neuron activations, computations can be skipped where inputs do not change from the previous update.
Because each neuron that is not updated will save fetches of entire columns of several weight matrices, efficiently determining which neurons need to be updated offers significant speedups.

The rest of this paper is organized as follows.  Sec.~\ref{sec:dnformulation} introduces the delta network concept in terms of the basic matrix-vector operations.
Sec.~\ref{sec:dngru} concretely formulates it for a GRU RNN.
{Sec.~\ref{sec:approx} proposes a method using a finite threshold for the deltas that suppresses the accumulation of the transient approximation error.}
Sec.~\ref{sec:methods} describes methods for optimally training a delta RNN. Sec.~\ref{sec:results} shows accuracy versus speedup for three examples.
Finally, Sec.~\ref{sec:conclusion} compares this work with other developments and summarizes the results.

\section{Delta Network Formulation}
\label{sec:dnformulation}

\begin{figure}[!t]
\centering
\includegraphics[width=0.48\textwidth]{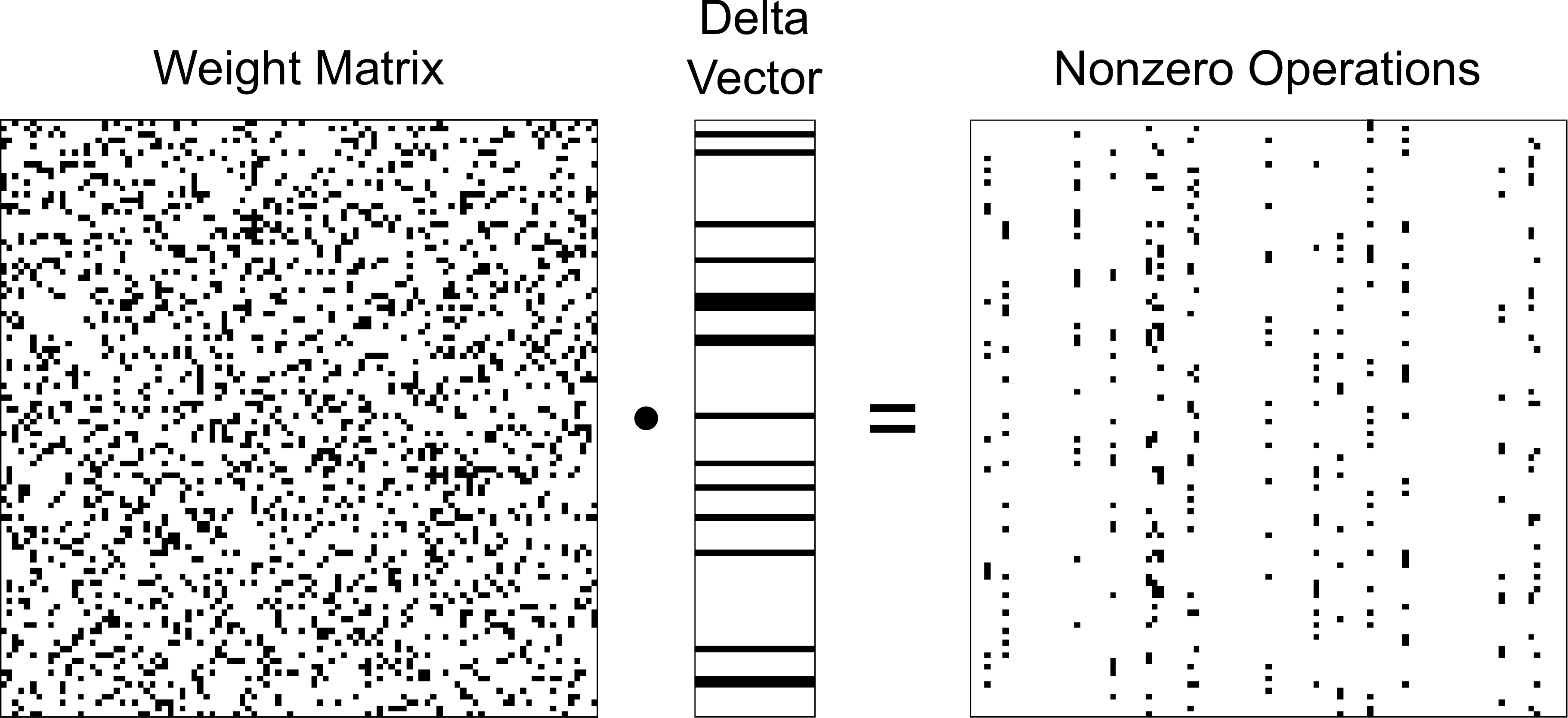}
\caption{Illustration of saved matrix-vector computation using delta networks with sparse delta vectors and weight matrices.  }
\label{fig:schematic_intuition}
\end{figure}

The purpose of a delta network is to transform a dense matrix-vector multiplication (for example, a weight matrix and a state vector) into a sparse matrix-vector multiplication followed by a full addition.  This transformation leads to savings on both operations (actual multiplications) and more importantly memory accesses (weight fetches).  Fig.~\ref{fig:schematic_intuition} illustrates the savings due to a sparse multiplicative vector. Zeros are shown with white, while non-zero matrix and vector values are shown in black.  Note the multiplicative effect of sparsity in the weight matrix and sparsity in the delta vector.  In this example, 20\% occupancy of the weight matrix and 20\% occupancy of the $\Delta$ vector requires fetching and computing only 4\% of the original operations.

To illustrate this methodology, consider a general matrix-vector multiplication of the form
\begin{IEEEeqnarray}{c}
r = W x
\end{IEEEeqnarray}
that uses $n^2$ compute operations\footnote{In this paper, a ``compute'' operation is either a multiply, an add, or a multiply-accumulate. The costs of these operations are similar, particularly when compared to the cost of an off-chip memory operation. See \cite{horowitz2014} for a simple comparison of energy costs of compute and memory operations}, $n^2 + n$ reads and $n$ writes for a $W$ matrix of size $n \times n$ and a vector $x$ of size $n$. Now consider multiple matrix-vector multiplications for a long input vector sequence $x_t$ indexed by $t = 1, 2, \ldots, n$. The corresponding result $r_t$ can be calculated recursively with:
\begin{IEEEeqnarray}{c}
r_t = W \Delta + r_{t-1},
\label{eq:delta_concept}
\end{IEEEeqnarray}
where $\Delta = x_t - x_{t-1}$ and $r_{t-1}$ is the result obtained from the previous calculation; if stored, the compute cost of $r_{t-1}$ is zero as it can be fetched from the previous timestep. Trivially, $x_0 = 0$ and $r_0 = 0$.  It is clear that
\begin{IEEEeqnarray}{cll}
r_t &= W (x_t - x_{t-1}) + r_{t-1}\\
    &= W (x_t - x_{t-1}) + W (x_{t-1} - x_{t-2}) + \ldots + r_{0}\\
    &= W x_t
\end{IEEEeqnarray}
Thus this formulation, which uses the difference between two subsequent steps and referred to as the {\it{delta network}} formulation, can be seen to produce exactly the same result as the original matrix-vector multiplication.

\subsection{Theoretical Cost Calculation}
\label{sec:costs}
The $\Delta$ from \eqref{eq:delta_concept} results in a reduction of computes and memory accesses of the weight matrix if $\Delta$ is a sparse vector.  To illustrate this, begin by defining $o_c$ to be the {\it{occupancy}} of a vector if a ratio $o_c$ of the vector elements are nonzero.  Consider the compute cost for $r_t$; it consists of the sum of the cost of calculating $\Delta$ (requiring $n$ operations for a vector of size $n$), the cost of adding in the stored previous result $r_{t-1}$ ($n$ operations),
and the cost of the sparse matrix multiply $W \Delta$ ($o_c \cdot n^2$ operations for an $n \times n$ weight matrix and a sparse $\Delta$ vector of occupancy ratio $o_c$).
Similarly, the memory cost for calculating  $r_t$ requires fetching
$o_c \cdot n^2$ weights for $W$, $2n$ values for $\Delta$, $n$ values for $r_{t-1}$ and writing out the $n$ values of the result.


Overall, the compute cost for the standard formulation ($C_{\rm comp, dense}$) and the new delta formulation ($C_{\rm comp, sparse}$) will be:
\begin{IEEEeqnarray}{rl}
C_{\rm comp, dense}  & = n^2\\
C_{\rm comp, sparse} & = o_c \cdot n^2 + 2n
\end{IEEEeqnarray}
while the memory access costs for both the standard ($C_{\rm mem, dense}$) and delta networks ($C_{\rm mem, sparse}$) can be seen from inspection as:
\begin{IEEEeqnarray}{rl}
C_{\rm mem, dense}  & = n^2 + n\\
C_{\rm mem, sparse} & = o_c \cdot n^2 + 4n
\end{IEEEeqnarray}
Thus, the arithmetic intensity (ratio of arithmetic to memory access costs) as $n\to\infty$ is 1 for both the standard and delta network methods.  This means that for both methods of calculating $r_t$, every arithmetic operation requires a memory access, unfortunately placing computational accelerators at a disadvantage.  However, if a sparse occupancy $o_c$ of $\Delta$ is assumed, then the decrease in computes and memory accesses due to storing the previous state will result in a
speedup of:
\begin{IEEEeqnarray}{rl}
\label{eq:speedup_occupancy}
C_{\rm dense}/C_{\rm sparse}  \approx n^2 / (n^2 o_c) = (1/o_c)
\end{IEEEeqnarray}
For example, if $o_c=10\%$, then the theoretical speedup will be 10X.

Note that this speedup is determined by the occupancy in each computed $\Delta = x_t - x_{t-1}$, implying that this sparsity is determined by the data stream.  Specifically, the regularity with which values stay exactly the same between $x_t$ and $x_{t-1}$, or as demonstrated later, within a certain absolute value called the threshold, determines the speedup.  In a neural network, the vector $x$ can represents inputs, intermediate activation values, or outputs of RNNs.  If $x$ changes slowly between subsequent timesteps then the input values $x_t$ and $x_{t-1}$ will be highly redundant, leading to a low occupancy $o_c$ and a correspondingly increased speedup.

\section{delta network GRUs}
\label{sec:dngru}
In GRUs, the matrix-vector multiplication operation that can be replaced with a delta network operation appears several times, shown in bold below. This GRU formulation is from \cite{chung2014empirical}:
\begin{IEEEeqnarray}{cl}
\label{eq:stdgru1}
r_t & = \sigma_r(\mathbf{W_{xr} x_t } + \mathbf{W_{hr} h_{t - 1}} + b_r)\\
u_t & = \sigma_u(\mathbf{W_{xu} x_t } + \mathbf{W_{hu} h_{t - 1}} + b_u)\\
c_t & = \sigma_c(\mathbf{W_{xc} x_t } + r_t \odot (\mathbf{W_{hc}} h_{t - 1}) + b_c)\\
h_t & = (1 - u_t) \odot h_{t - 1} + u_t \odot c_t
\label{eq:stdgru2}
\end{IEEEeqnarray}
Here $r$, $u$, $c$ and $h$ are reset and update gates, candidate activation, and activation vectors, respectively, typically a few hundred elements long. The $\sigma$ functions are nonlinear logistic sigmoids that saturate at 0 and 1. The $\odot$ signifies element-wise multiplication. Each term in bold can be replaced with the delta update defined in (\ref{eq:delta_concept}), 
forming:
\begin{IEEEeqnarray}{rCl}
\Delta_x & = & x_t - x_{t-1}\\
\Delta_h & = & h_{t - 1} - h_{t - 2}\\
r_t & = & \sigma_r(W_{xr} \Delta_x + z_{xr} + W_{hr} \Delta_h + z_{hr}  + b_r)\\
u_t & = & \sigma_u(W_{xu} \Delta_x + z_{xu} + W_{hu} \Delta_h + z_{hu}  + b_u)\\
c_t & = & \sigma_c(W_{xc} \Delta_x + z_{xc} + r_t \odot (W_{hc} \Delta_h + z_{hc})  + b_c)\\
h_t & = & (1 - u_t) \odot h_{t - 1} + u_t \odot c_t
\end{IEEEeqnarray}
where the values $z_{xr}$, $z_{xu}$, $z_{xc}, z_{hr}$, $z_{hu}$, $z_{hc}$ are recursively defined as the the stored result of the previous computation for the input or hidden state, i.e.:
\begin{IEEEeqnarray}{rCCCl}
z_{xr} & := & z_{xr, t-1} & = & W_{xr} (x_{t-1} - x_{t-2}) + z_{xr, t-2}
\end{IEEEeqnarray}
The above operation can be applied for the other five values $z_{xu}$, $z_{xc}, z_{hr}$, $z_{hu}$, $z_{hc}$.  The initial condition at time $x_0$ is $z_0 := 0$. 
Also, as can be seen from the equations above, many of the additive terms, including the stored full-rank pre-activation states as well as the biases, can be merged into single values resulting into four stored memory values ($M_{r}$, $M_{u}$, $M_{xc}$, and $M_{hr}$) for the three gates:
\begin{IEEEeqnarray}{rCCCl}
M_{t-1} & := & z_{x, t-1} + z_{h, t-1} + b
\end{IEEEeqnarray}
Finally, in accordance with the above definitions of the initial state, the memories $M$ are initialized at their corresponding biases, i.e., $M_{r, 0}=b_r$, $M_{u, 0}=b_u$, $M_{xc, 0}=b_c$, and $M_{hr, 0}=0$, resulting in the following full formulation of the delta network GRU:
\begin{IEEEeqnarray}{rCl}
\label{eq:dngru1}
\Delta_x & = & x_t - x_{t-1}\\
\Delta_h & = & h_{t - 1} - h_{t - 2}\\
M_{r, t} & := & W_{xr} \Delta_x  + W_{hr} \Delta_h + M_{r, t-1} \\
M_{u, t} & := & W_{xu} \Delta_x  + W_{hu} \Delta_h + M_{u, t-1} \\
M_{xc, t} & := & W_{xc} \Delta_x + M_{xc, t-1} \\
M_{hc, t} & := & W_{hc} \Delta_h  + M_{hc, t-1} \\
r_t & = & \sigma_r(M_{r,t})\\
u_t & = & \sigma_u(M_{u,t})\\
c_t & = & \sigma_c(M_{xc,t} + r_t \odot M_{hc,t} )\\
h_t & = & (1 - u_t) \odot h_{t - 1} + u_t \odot c_t
\label{eq:dngru2}
\end{IEEEeqnarray}

\section{Approximate Calculations in Delta Networks}
\label{sec:approx}
Note that the formulations described in Secs.~\ref{sec:dnformulation} and~\ref{sec:dngru} are designed to give precisely the same answer as the original computation in the network.  However, a more aggressive approach can be taken in the update, inspired by recent studies that have shown the possibility of greatly reducing weight precision in neural networks without giving up accuracy \cite{stromatiasneil2015robustness,courbariaux2014low}.
Instead of skipping a vector-multiplication computation if a change in the activation $\Delta = 0$, a vector-multiplication can be skipped if a value of $\Delta$ is smaller than the threshold (i.e $|\Delta_{i,t}| < \Theta$, where $\Theta$ is a chosen threshold value for a state $i$ at time $t$).  That is, if a neuron's hidden-state $M$ activation has changed by less than $\Theta$ since it was last memorized,
the neuron output will not be propagated, i.e., its $\Delta$ value is set to zero for that update. Using this threshold, the network will not produce precisely the same result at each update, but will produce a result which is approximately correct.  Moreover, the using a threshold substantially increases activation sparsity.

Importantly, if a non-zero threshold is used with a naive delta change propagation, errors can accumulate over multiple time steps through state drift.
For example, if the input value $x_t$ increases by nearly $\Theta$ on every time step, no change will ever be triggered despite an accumulated significant change in activation, causing a large drift in error.
Therefore, in our implementation, the memory records the last value causing an above-threshold change, not the difference since the last time step.

More formally, we introduce the states $\hat{x}_{i, t-1}$ and $\hat{h}_{j, t-1}$.  These states store the $i-$th input and the hidden state of the $j-$th neurons, respectively, at their last {\it{change}}.  The current input $x_{i, t}$ and state $h_{j, t}$ will be compared against these values to determine the $\Delta$.
Then the $\hat{x}_{i, t-1}$ and $\hat{h}_{j, t-1}$ values will only be updated if the threshold is crossed:
\begin{IEEEeqnarray}{rll}
\label{eq:dnthresh1}
\hat{x}_{i, t-1} & = &
\begin{cases}
    x_{i,t-1} & \text{if}\ |x_{i, t} - \hat{x}_{i, t-1} | > \Theta\\
    \hat{x}_{i, t-2} & \text{otherwise} \\
\end{cases} \\
\Delta{x_{i,t}} & = &
\begin{cases}
    x_{i, t} - \hat{x}_{i, t-1} & \text{if}\ |x_{i, t} - \hat{x}_{i, t-1} | > \Theta\\
    0 & \text{otherwise} \\
\end{cases} \\
\hat{h}_{j, t-1} & = &
\begin{cases}
    h_{j, t-1} & \text{if}\ |h_{j, t} - \hat{h}_{j, t-1} | > \Theta\\
    \hat{h}_{j, t-2} & \text{otherwise} \\
\end{cases} \\
\Delta{h_{j, t}} & = &
\begin{cases}
    h_{j, t} - \hat{h}_{j, t-1} & \text{if}\ |h_{j, t} - \hat{h}_{j, t-1} | > \Theta\\
    0 & \text{otherwise} \\
\end{cases}
\label{eq:dnthresh2}
\end{IEEEeqnarray}
That is, when calculating the input delta vector $\Delta{x_{i, t}}$ comprised of each element $i$ at time $t$, the difference between two values are used: the current value of the input $x_{i,t}$, and the value the last time the delta vector was nonzero $\hat{x}_{i, t-1}$.
Furthermore, if the delta change is under the threshold $\Theta$, then the delta change is set to zero, producing a small approximation error that will be corrected when a sufficiently large change produces a nonzero update.
The same formulation is used for the hidden state delta vector $\Delta{h_{j, t}}$.

\section{Methods for Reducing Approximation Error and Increasing Speedup}
\label{sec:methods}
This section presents training methods, constraints, and optimization schemes that yield faster and more accurate delta networks.

\subsection{Rounding Network Activations}\label{sec:roundmethod}
The thresholded delta network computation described in Sec.~\ref{sec:approx} performs a rounding function similar to a rounding of the partially-computed state, since small changes are rounded to zero while large changes are propagated.
Since many previous investigations have demonstrated methods to train networks to be robust against small rounding errors by rounding during training, one method that could increase accuracy is to perform activation rounding.
Then, using the techniques outlined in \cite{courbariaux2014low,stromatiasneil2015robustness}, a network can be successfully trained so that it is robust to these small rounding errors.  Furthermore, low-precision computation and low-precision parameters can further reduce power consumption and improve the efficiency of the network for dedicated hardware implementations.

As explored in previous studies, a low-resolution activation $\theta_L$ in signed fixed-point format $Qm.f$ with $m$ integer bits and $f$ fractional bits can be produced from a high-resolution activation $\theta$ by using a deterministic and gradient-preserving rounding:
\begin{IEEEeqnarray}{cl}
\theta_L = \text{round}(2^f \cdot \theta) \cdot 2^{-f}
\end{IEEEeqnarray}
with $2^f \cdot \theta$ clipped to a range $[-2^{m+f-1}, 2^{m+f-1}]$.  Thus, the output error cost will incorporate the errors due to small rounding approximations, and the process of stochastic gradient descent used to increase the accuracy will learn to avoid these errors through exposure during training.

\subsection{Adding Gaussian Noise to Network Activations}\label{sec:gaussmethod}
Once thresholding has been introduced, the network must be robust to the non-propagation of small changes, while large changes should be considered important.  Another way to provide robustness against small changes is to add Gaussian noise $\eta$ to terms that will have a thresholded delta activation:
\begin{IEEEeqnarray}{lll}
r_t & = \sigma_r((x_t + \eta_{x}) W_{xr} + (h_{t - 1}+\eta_{h}) W_{hr} + b_r) \\
u_t & = \sigma_u((x_t + \eta_{x}) W_{xu} + (h_{t - 1}+\eta_{h}) W_{hu} + b_u)\\
c_t & = \sigma_c((x_t + \eta_{x}) W_{xc} + r_t \odot ((h_{t - 1}+\eta_{h}) W_{hc}) + b_c)\\
h_t & = (1 - u_t) \odot h_{t - 1} + u_t \odot c_t
\end{IEEEeqnarray}
where $\eta  \sim \mathcal{N}(\mu, \sigma)$. That is, $\eta$ is a vector of samples drawn from the Gaussian distribution with mean $\mu$ and variance $\sigma$,
and $\eta \in \{\eta_{x}, \eta_{h}\}$. Each element of these vectors is drawn independently.  Typically, the value $\mu$ is set to $0$ so that the expectation is unbiased, e.g., $\mathbf{E}[x_t + \eta_{x}] = \mathbf{E}[x_t]$.

As a result, the Gaussian noise should prevent the network from being sensitive to minor fluctuations, and increase its robustness to truncation errors.

\subsection{Training Directly on Delta Networks}
\label{sec:trainingdirectly}
However, injecting Gaussian noise at many points in the network computation is still not the same as the truncation operation performed by a thresholded delta network.  To best model that truncation, the network should be trained directly on the errors that arise from a delta network.  The resulting network will then be robust against the types of errors that a thresholding delta network typically makes.

More accurately, instead of training on the original GRU equations Eq.~\ref{eq:stdgru1}--\ref{eq:stdgru2}, the state is updated using the delta network model described in Eq.~\ref{eq:dngru1}--\ref{eq:dnthresh2}.  This change should incur no accuracy loss between train accuracy and test accuracy, but the model may yet have more difficulty during the training if the model proves harder to optimize and possibly result in an overall lower accuracy level.

\subsection{Considering Additional Speedup from Weight Sparsity}
\label{sec:weightsparsity}
Furthermore, the speedup from using a delta network so far has been considered to only arise from the sparse delta vectors that allow skipping columns of the weight matrices. However, the amount of sparsity in the weight matrices of deep networks after training also can affect the savings in the computational cost and the speedup.
Studies such as in~\cite{ott2016recurrent} show that in trained low-precision networks, the weight matrices can be quite sparse. For example, in a ternary or 3-bit weight network the weight matrix sparsity can exceed 80\% for small RNNs.
Since every nonzero input vector element is multiplied by a column of the weight matrix, this computation can be skipped if the weight value is zero.
That is, the zeros in the weight matrix act multiplicatively with the delta vector to produce even fewer necessary multiply-accumulates, as illustrated above in Fig.~\ref{fig:schematic_intuition}.  The calculation of the matrix-vector product then costs:
\begin{IEEEeqnarray}{rl}
C_{\rm comp, sparse} & = o_m \cdot o_c \cdot n^2 + 2n \\
C_{\rm mem, sparse} & = o_m \cdot o_c \cdot n^2 + 4n
\end{IEEEeqnarray}
for a weight matrix with occupancy $o_m$.  By comparison to Eq.~\ref{eq:speedup_occupancy}, the system can achieve a theoretical speedup of $1/(o_m \cdot o_c)$.  That is, by compressing the weight matrix and only fetching nonzero weight elements that combine with the nonzero state vector, a higher speedup can be obtained without degrading the accuracy.

\subsection{Incurring Sparsity Cost on Changes in Activation}
\label{sec:sparsitychangetraining}
Finally, if the network is trained using the delta network model, a cost can be associated with the delta terms and added into the overall cost.  In a batch of input samples, the $L_1$ norm for $\Delta_h$ can be calculated as the mean absolute delta changes, and this norm can be scaled by a weighting factor $\beta$.  This $L_{\rm sparse}$ cost can then be additively incorporated into the standard loss function.  That is:
\begin{IEEEeqnarray}{cl}
    \mathcal{L}_{\rm sparse} = \beta ||\Delta{h}||_1
\end{IEEEeqnarray}
Here the $L_1$ norm is used to encourage sparse values in $\Delta{h}$, so that fewer delta updates are required.

\section{Results}
\label{sec:results}
This section presents the results showing the trade-off between compute savings and accuracy loss from RNNs trained on the TIDIGITS digit recognition benchmark.  Furthermore, it also demonstrates that the results found on small datasets also appear in the much larger Wall Street Journal speech recognition benchmark. The final example is for a CNN-RNN stack trained on end-to-end steering control using a recent driving dataset.
The fixed-point $Q3.4$ (i.e $m=3$ and $f=4$) format was used for network activation values in all speech experiments except the ``Original'' RNN line for TIDIGITS in Fig. 4, which was trained in floating-point representation. The driving dataset in \ref{sec:driving} used Q2.5 activation.  The networks were trained with Lasagne \cite{sander_dieleman_2015_27878} powered by Theano \cite{bergstra2010theano}.  The training time on an Nvidia GTX980 Ti GPU is reported to indicate training difficulty, per discussions in the deep learning symposium at NIPS 2016.

\subsection{TIDIGITS dataset}
\label{sec:tidigits}
The TIDIGITS dataset \cite{leonard1993tidigits} was used as an initial evaluation task for the methods introduced in Sec.~\ref{sec:methods}.  Single digits (``oh'' and zero through nine) from this database, with a total of 2464 digits in the training set and 2486 digits in the test set, were transformed in the standard way \cite{neil2016effective} to produce a 39-dimensional Mel-Frequency Cepstral Coefficient (MFCC) feature vector using a 25\,ms window, 10\,ms frame shift, and 20 filter bank channels.  The labels for ``oh'' and ``zero" were collapsed to a single label.  Training time is approximately 8 minutes for 150 epochs of training per experiment.

\begin{figure}[!t]
\centering
\includegraphics[width=0.48\textwidth]{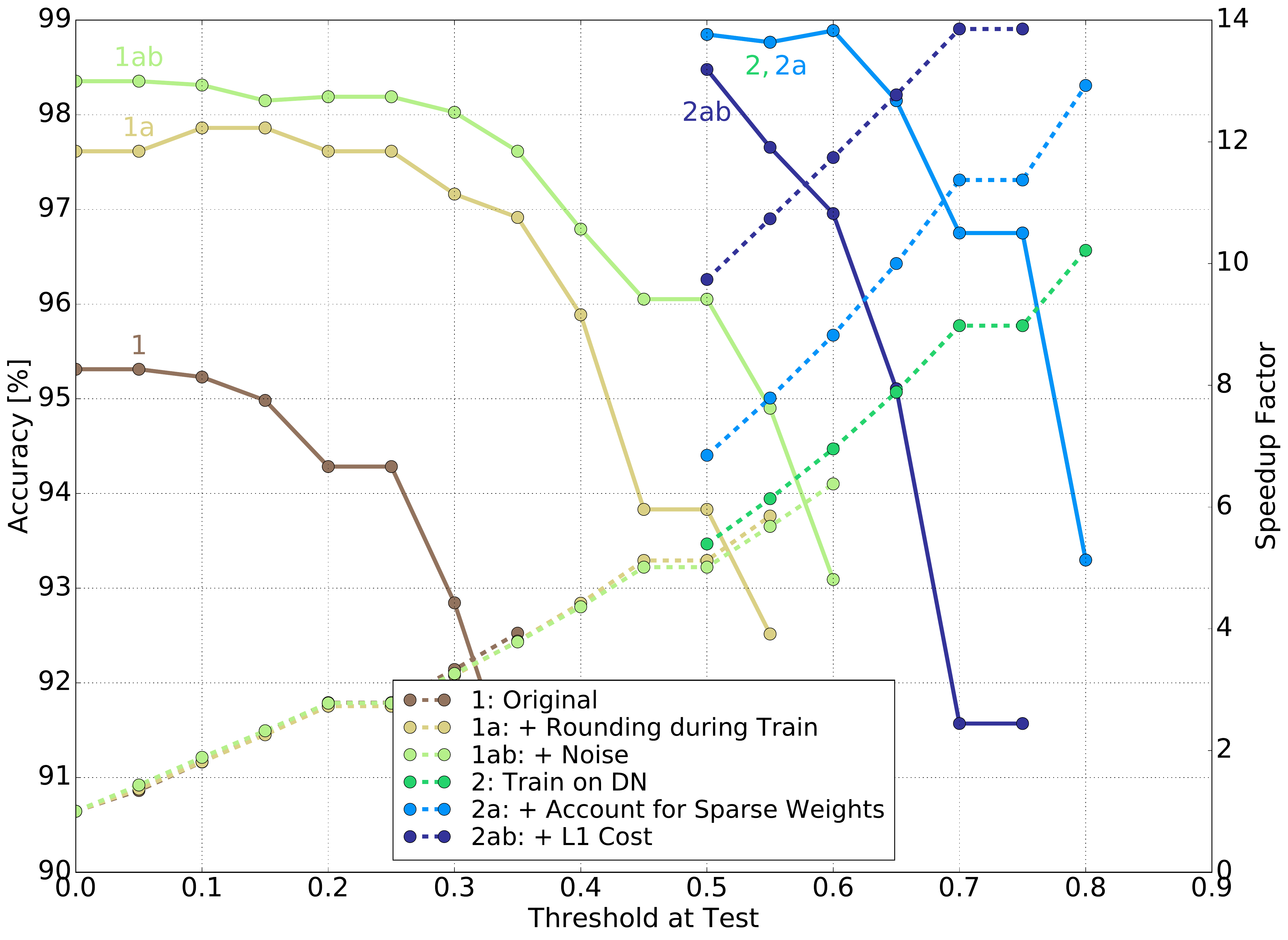}
\caption{Test accuracy results from standard GRUs run as delta networks after training (curves {\bf{1}}, {\bf{1a}}, and {\bf{1ab}}) and those trained as delta networks (curves {\bf{2}}, {\bf{2a}}, and {\bf{2ab}}) under different constraints on the TIDIGITS dataset. The delta networks are trained for $\Theta=0.5$. Note that the methods are combined, hence the naming scheme.  Additionally, the accuracy curve for {\bf{2}} is hidden by the curve {\bf{2a}}, since both achieve the same accuracy and only differ in speedup metric.  }
\label{fig:dcn_gru_tidigits_constraint_biplot}
\end{figure}

\begin{figure}[!t]
\centering
\includegraphics[width=0.48\textwidth]{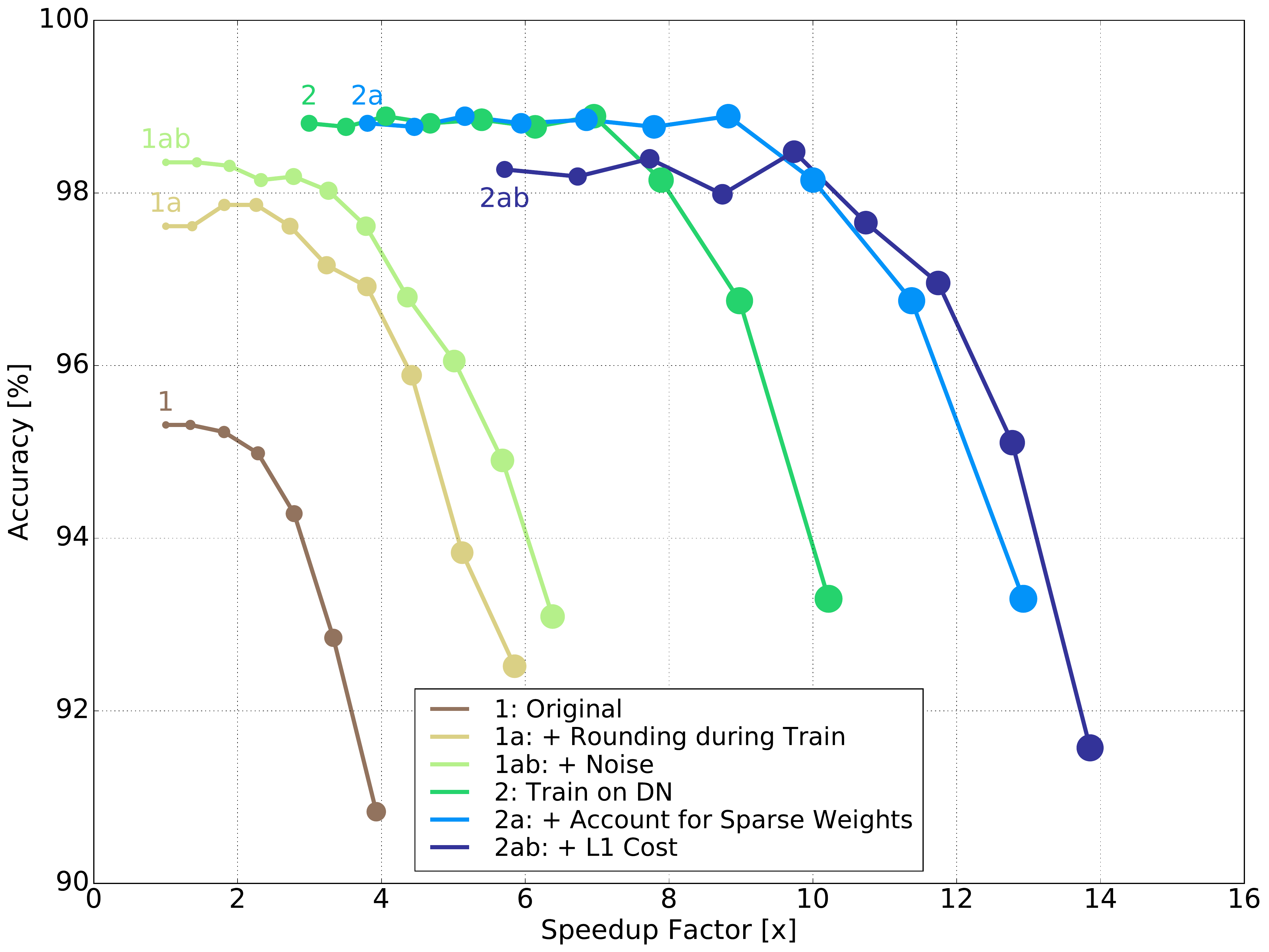}
\caption{Accuracy-speedup tradeoff by adjusting threshold for TIDIGITS dataset.  By increasing the threshold (indicated by sample point size), greater speedups can be obtained at greater losses of accuracy.  For networks trained as delta networks, the training threshold is the first (leftmost) point in the line point sequence.}
\label{fig:dcn_gru_tidigits_contour}
\end{figure}

The results of applying the methods introduced in Sec.~\ref{sec:methods} can be found in Fig.~\ref{fig:dcn_gru_tidigits_constraint_biplot}. There are two quantities measured: the change in the number of memory fetches, and the accuracy as a function of the threshold $\Theta$.  Fig.~\ref{fig:dcn_gru_tidigits_contour} shows the same results, but removes the threshold axis to allow easier comparison among the different training methods.

First, a standard GRU RNN was trained, achieving 96.59\% accuracy on the TIDIGITS task without data augmentation and regularization. This network consists of a layer of 200 GRU units connected to a layer of 200 fully-connected units and finally to a classification layer for the 10 digits.  This network was then subsequently tested using the delta network GRU formulation given in Sec.~\ref{sec:dngru}.

The standard RNN run as a delta network (``Original'') achieves 95\% accuracy (a drop from zero delta threshold accuracy of 96\%) with a speedup factor of about 2.2X.  That is, only approximately 45\% of the computes or fetches are needed in achieving this accuracy.
By adding the rounding constraint during training (``+~Rounding during Training''), the accuracy is nearly 97\% with an increase to a 3.9X speedup. By incorporating Gaussian noise (``+~Noise''), the accuracy can be boosted even further to about 97\% with a 4.2X speedup.  Essentially, these methods added generalization robustness to the original GRU, while preventing small changes from influencing the output of the network.  These techniques allow a higher threshold to be used while maintaining the same accuracy, which results in a decrease of memory fetches and a corresponding speedup.

Furthermore, training on the delta network itself (``Train on DN'') allows a considerable speedup, achieving 98\% accuracy with a 8X speedup.  Accounting for the effect of pre-existing sparsity in the weight matrix (``+~Account for Sparse Weights'') increases the speedup to 10X, without affecting the accuracy (as it is the same network).  Finally, incorporating an L1 cost on network changes in addition to training on the delta network model (``+~L1 cost'') achieves 97\% accuracy while boosting speedup to 11.9X. Adding in the sparseness cost on network changes decreases the accuracy slightly, since the loss minimization must find a tradeoff between both error and delta activation instead of considering error alone.  However, using the L1 loss can offer a significant additional speedup while retaining an accuracy increase over the original GRU network.

Finally, Fig.~\ref{fig:dcn_gru_tidigits_contour} also demonstrates the primary advantage given by each algorithm; an increase in generalization robustness manifests as an overall upward shift in accuracy, while an increase in sparsity manifests as a rightward shift in speedup.  Methods 1, 1a, and 1b generally increase generalization robustness while only modestly influencing the sparsity.  Method 2 greatly increases both, while method 2a only increases sparsity, and finally method 2ab slightly decreases accuracy but offers the highest speedup.

\subsection{Wall Street Journal dataset}
\label{sec:wsj}
 While the gains seen on TIDIGITs are significant, the delta network methodology was applied to an RNN trained on a larger dataset to determine whether it could produce the same gains.  Here, the Wall Street Journal dataset comprised of 81~hours of transcribed speech, as described in \cite{braun2016curriculum}.  Similar to that study, the first 4 layers of the network consisted of bidirectional GRU units with 320 units in each direction. Training time for each experiment was about 120h.

\begin{figure}[!t]
\centering
\includegraphics[width=0.48\textwidth]{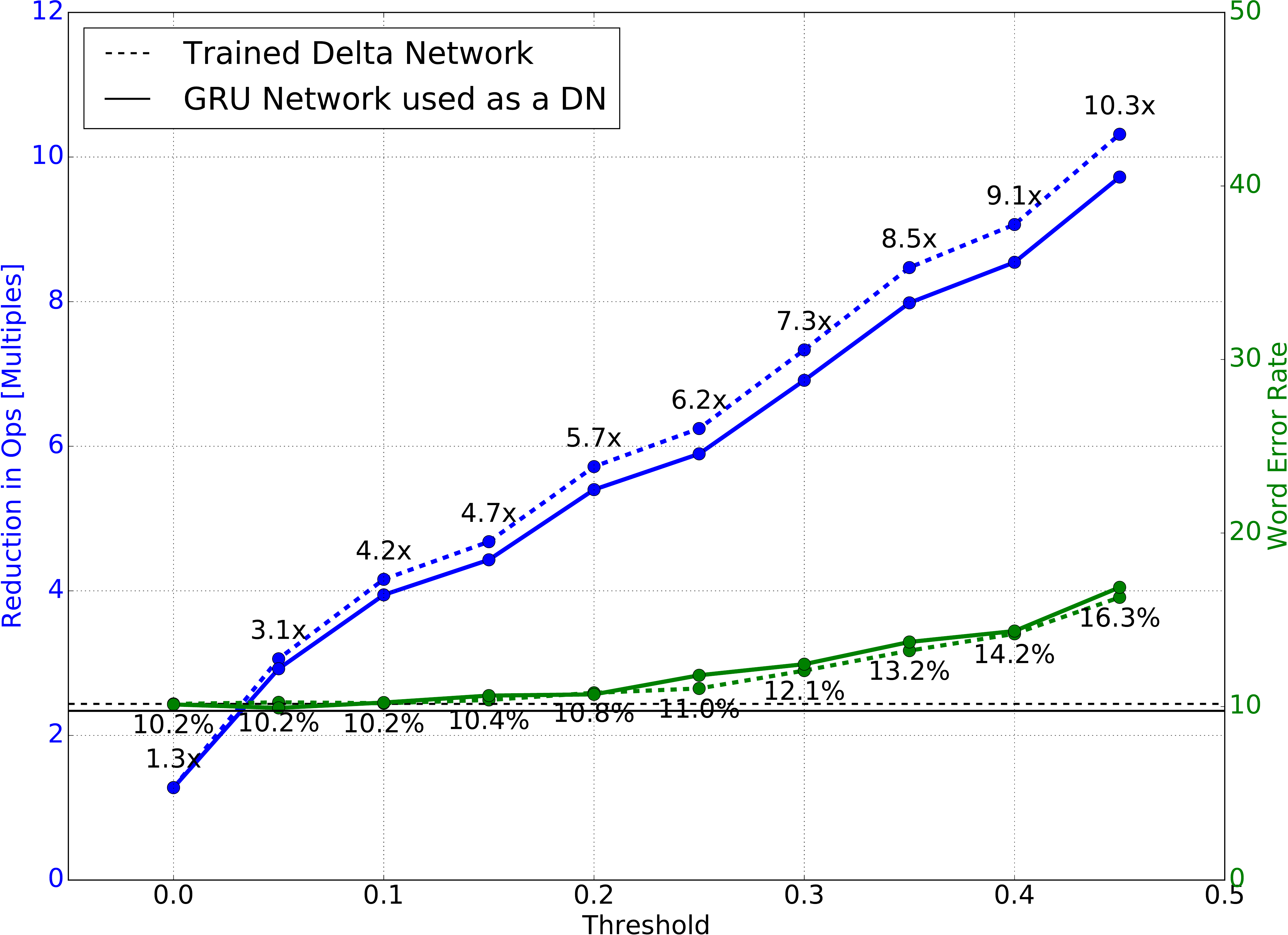}
\caption{Accuracy and speedup tradeoffs on the Wall Street Journal dataset.  The solid lines show results from an existing deep RNN trained on this dataset but
run as a delta network.  The dashed lines show results from a network trained as a delta network using $\Theta=0.2$ during training.
The horizontal lines indicate the non-delta network accuracy level; similarly, the solid and dashed horizontal lines indicate the accuracy of the normal network and the DN network prior to rounding, respectively.}
\label{fig:WSJ}
\end{figure}

Fig.~\ref{fig:WSJ} presents results on the achieved word error rate (WER) and speedup on this dataset for two cases: First, running an existing speech transcription RNN as a delta network
(results shown as solid curves labeled ``RNN used as a DN''), and second, a network trained as a delta network with results shown as the dashed curves ``Trained Delta Network''.  The speedup here accounts for weight matrix sparsity as described in Sec.~\ref{sec:weightsparsity} .

Surprisingly, the existing highly trained network already shows significant speedup without loss of accuracy as the threshold, $\Theta$, is increased: At $\Theta=0.2$, the speedup is about 5.5 with a WER of 10.8\% compared with the WER of 10.2\% at $\Theta=0$.
However, training the RNN to run as a delta network yields a network that achieves a slightly higher 5.7X speedup with the same WER. For this large, multilayer RNN that processes complex and constantly-changing speech data, even the conventionally-trained RNN run as a delta network can provide greater than 5X speedup with only a 5\% increase in the WER.

\subsection{Comma.ai Driving DataSet}
\label{sec:driving}
\begin{figure}[!t]
\centering
\includegraphics[width=0.48\textwidth]{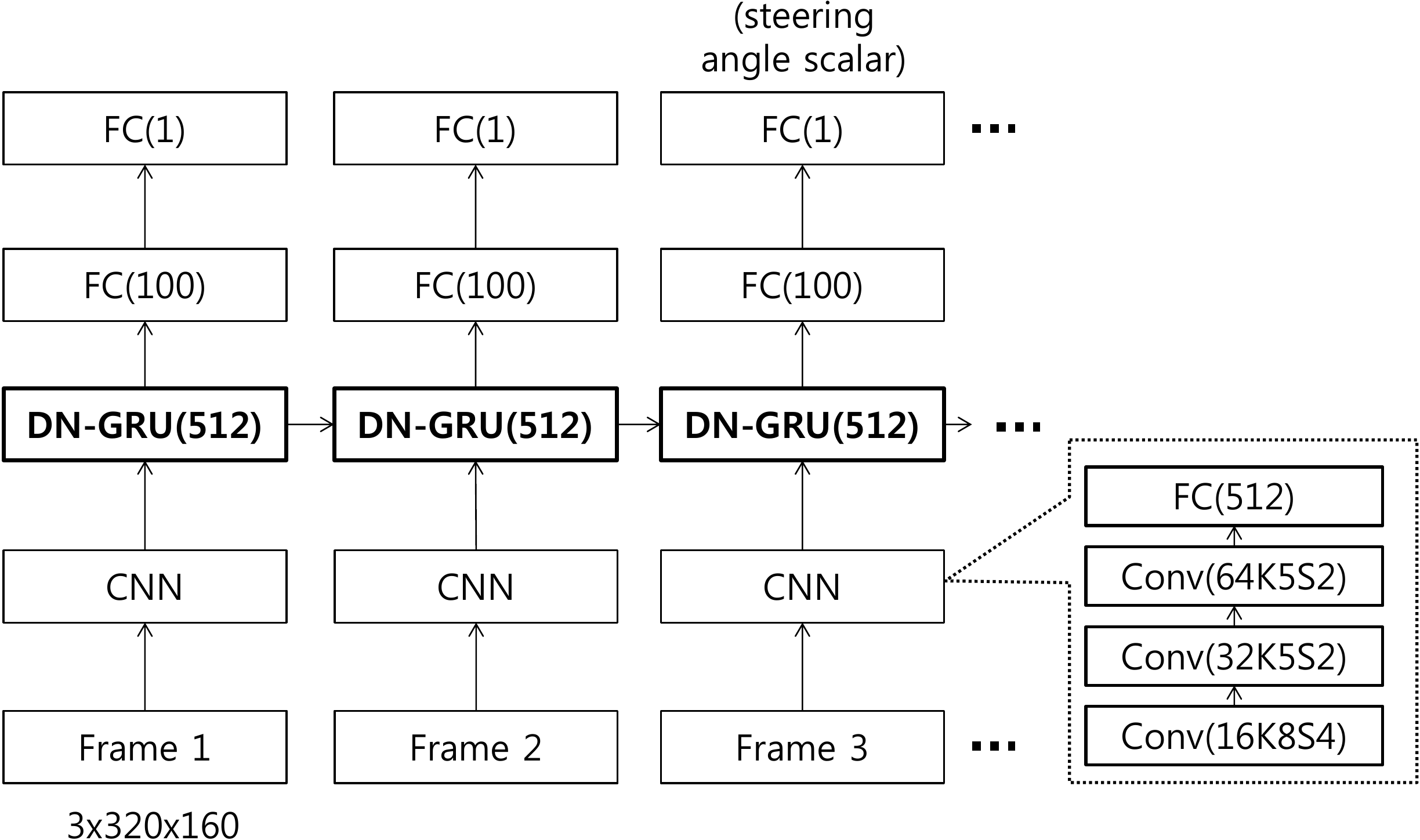}
\caption{Network architecture for steering angle prediction. The CNN feature detector consists of three convolution layers (Conv) and a fully-connected layer (FC) with 512 units. Conv(64K5S2) represents a convolution layer with 64 feature maps and 5x5 kernel with stride 2. Visual features of each image frame are fed into the GRU-based RNN to predict steering angle.}
\label{fig:driving_net_arch}
\end{figure}

While speech applications are a common area of exploration for RNNs, driving scenarios are rapidly emerging as another area of focused research.  Here, the delta network model was applied to determine the gains of exploiting the redundancy of real-time video input.   The open driving dataset from comma.ai \cite{santana2016commaai} with 7.25~hours of driving data was used, with video data recorded at 20\,FPS from a camera mounted on the windshield.
The network is trained to predict the steering angle from the visual scene similar to \cite{hempel2016driving, bojarski2016driving}. We followed the approach in \cite{hempel2016driving} by using an RNN on top of the CNN feature detector as shown in Fig. \ref{fig:driving_net_arch}.
The CNN feature detector has three convolution layers without pooling layers and a fully-connected layer with 512 units. During training, the CNN feature detector was pre-trained with an analog output unit to learn the recorded steering angle from randomly selected single frame images. Afterwards, the delta network RNN was added, and was trained by feeding sequences of the visual features from the CNN feature detector to learn sequences of the steering angle. Since Q2.5 format was used for the GRU layer activations, the GRU input and output vectors were normalized to match this range.

However, this raw dataset results in a few practical difficulties and requires data preprocessing. In particular, the driver's intention of changing lanes or taking a specific route at the fork of a road cannot be learned by training on this dataset, as no intent or route goal is provided (though could be addressed in the future using precise route planning and localization techniques).  Additionally, the ground truth of the steering angle is noisy due to the sometimes unpredictable behavior of the driver; for example, the driver occasionally idly adjusts the steering wheel when the car is at a stop.  As a a result, the recorded steering angle occasionally becomes uncorrelated with the direction of movement of the car. However, this issue can be addressed in a straightforward way by excluding the frames recorded during periods of low speed driving. Training time of the CNN feature detector was about 8 hours for 10k updates with the batch size of 200. Training of the RNN part took about 3 hours for 5k updates with the batch size of 32 samples consisting of 48 frames/sample.

\begin{figure}[!t]
\centering
\includegraphics[width=0.48\textwidth]{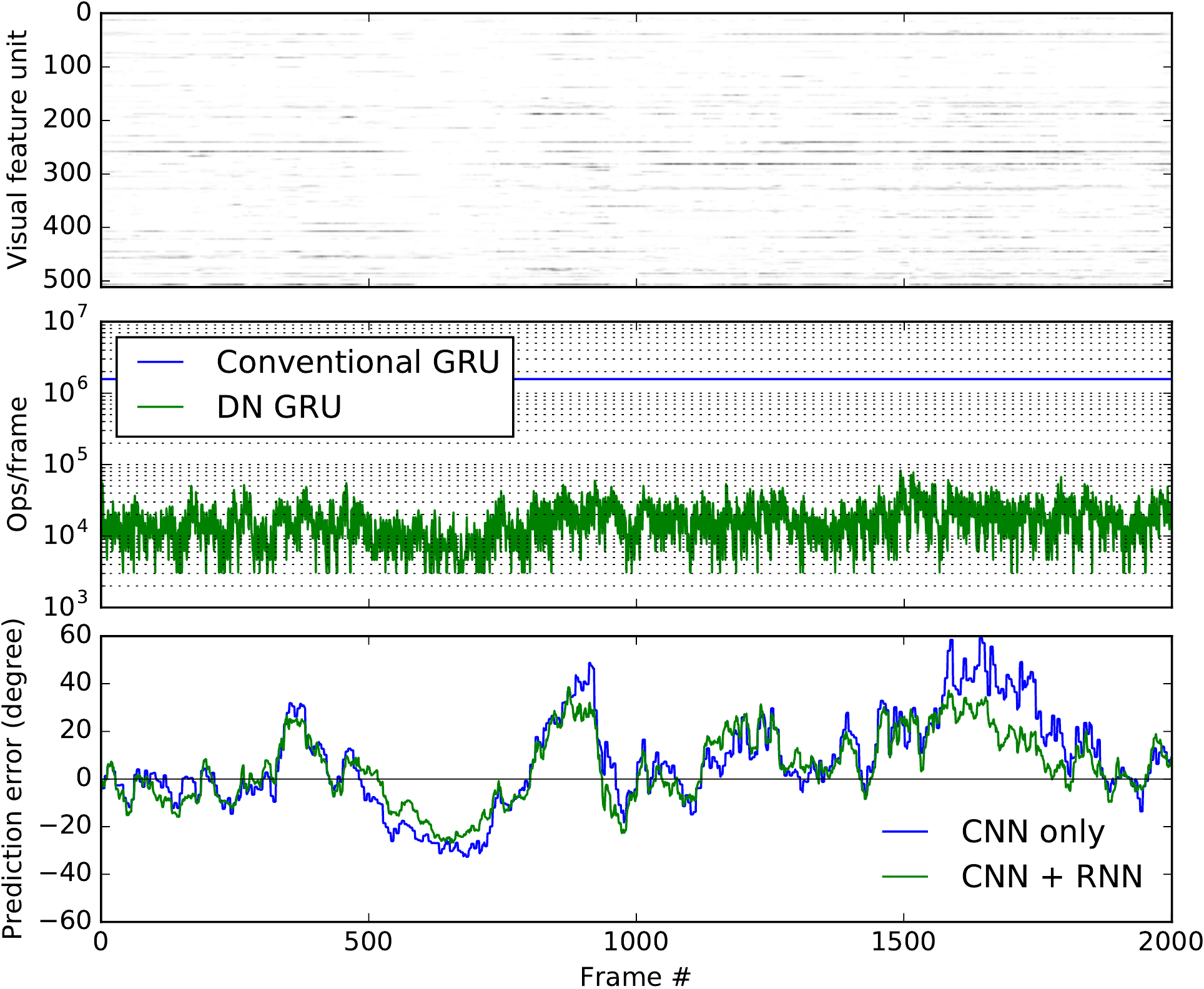}
\caption{Reduction of RNN compute cost in the steering angle prediction task on the comma.ai driving dataset.
The top figure shows the output of CNN feature detector. The middle figure shows the required \# of ops per frame for the delta network GRU layer (trained with $\Theta=0.1$) in comparison with the conventional GRU case.
A huge speedup is obtained because of the large temporal redundancy of the driving visual scenes. The bottom figure compares the prediction errors of CNN predictor and CNN+RNN predictor.
Note that the RNN slightly improves the steering angle prediction by using multiple frames. (See Fig.~\ref{fig:driving_speedup})}
\label{fig:driving}
\end{figure}

\begin{figure}[!t]
\centering
\includegraphics[width=0.48\textwidth]{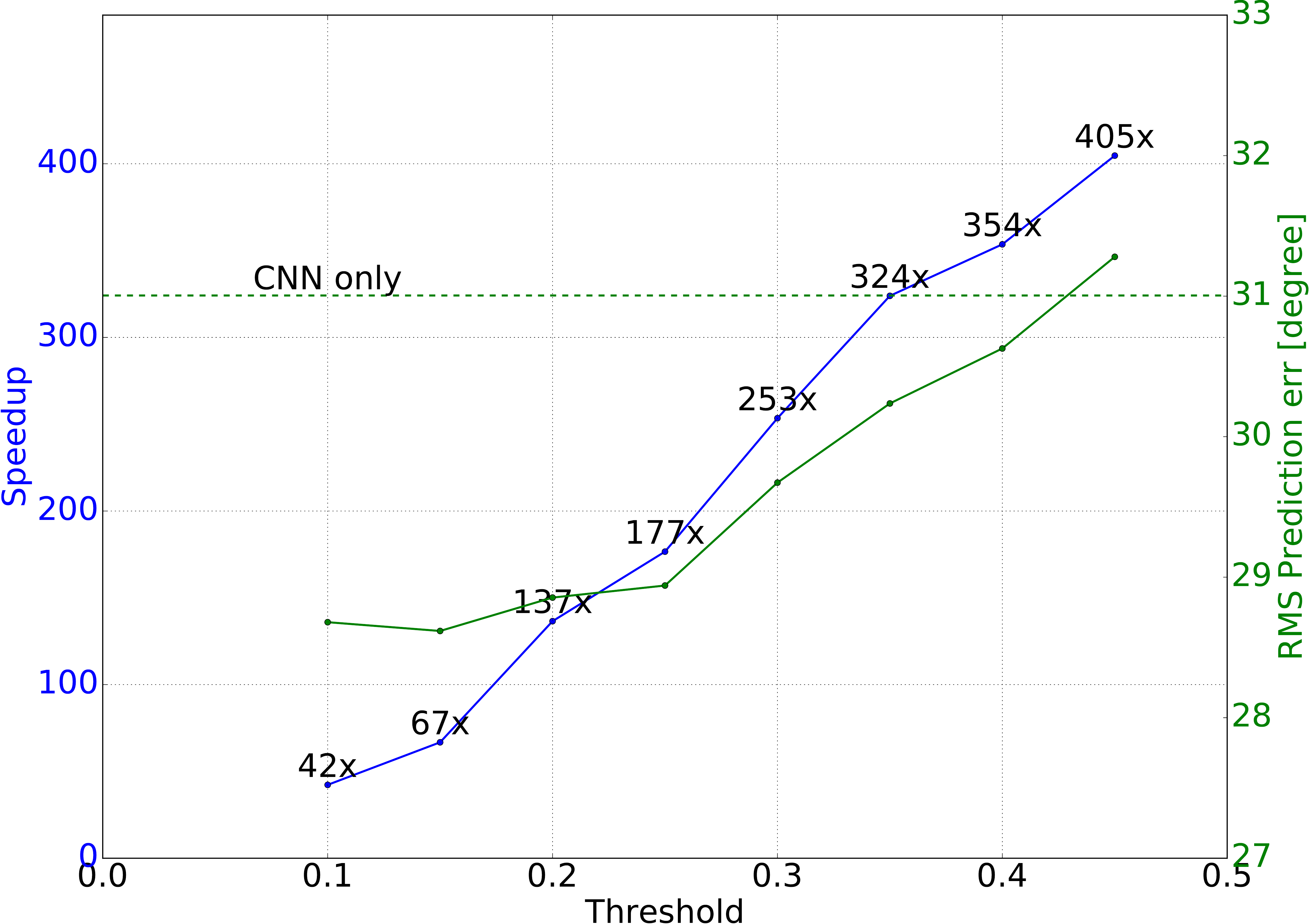}
\caption{Tradeoffs between prediction error and speedup of the GRU layer on the steering angle prediction of the comma.ai driving dataset. The result was obtained from 1000 samples with 48 consecutive frames sampled from the validation set. Speedup here does not include weight matrix sparsity. The network was trained with $\Theta=0.1$. A speedup of approximately 100X can be obtained without increasing the prediction error, using $\Theta$ between 0.1 and 0.25.}
\label{fig:driving_speedup}
\end{figure}

Fig.~\ref{fig:driving} shows the compute cost of the delta network GRU layer in comparison with a conventional GRU, in the steering angle prediction task on 2000 consecutive frames (100s) from the validation set. While the number of operations per frame remains constant for the conventional GRU layer, those for the delta network GRU layer varies dynamically depending on the change of visual features.
Since the output of the CNN feature detector is very stable over time as shown in the top figure, a huge speedup of about 100X (see Fig.~\ref{fig:driving_speedup})  is obtained by removing the large temporal redundancy of the visual features in driving data.

However, in this steering network, the computational cost of the CNN (about 37\,MOp/frame) dominates the RNN cost (about 1.58\,MOp/frame).  Thus the overall, system-level computational savings for this example is only about 4.2\%.  However, future applications will likely have efficient dedicated vision hardware or require a greater role for RNNs in processing numerous and complex data streams, which result in RNN models that consume a greater percentage of the overall energy/compute cost.  Even now, steering angle prediction already benefits from a delta network approach.

\section{Discussion and Conclusion}
\label{sec:conclusion}

Although the delta network concept can be applied to
other network architectures, as was shown in similar concurrent work for CNNs~\cite{oconnor2016sigmadelta},
in practice a larger benefit is seen in RNNs because
we already need to store all the intermediate activation values for the delta networks.
For example, the widely-used VGG19 CNN has 16M neuron states~\cite{chatfield2014return}. Employing the delta network approach for CNNs
 requires doubled memory access and a lot of additional memory space for neuron states.
Because the cost of external memory access is hundreds of times larger than that of arithmetic operations,
delta network CNNs seem impractical without new memory technology to address this issue.

In contrast, for RNNs, the number of weight parameters is much bigger than the number of activations.
The sparsity of the deltas allows large savings in power consumption by reducing the number of memory access for weight parameters.
CNNs do not have this advantage since the weight parameters are
shared by many units and their number is much smaller than the number of activations.
Whereas the work in~\cite{oconnor2016sigmadelta}
focuses on an optimization method for converting a pre-trained CNN into a Sigma-Delta network to reduce the compute cost,
our work shows that the delta networks can be
optimized in terms of accuracy and speedup by directly training the original network to run as a delta network.

Recurrent neural networks can be highly optimized due to the redundancy of their activations over time.
 When the use of this temporal redundancy is combined with robust training algorithms, this work demonstrates that
 speedups of 6X to 9X can be obtained with negligible accuracy loss in speech RNNs, and
 speedups of over 100X are possible in steering angle prediction RNNs.

\noindent{{\bf Acknowledgements:} We thank S.~Braun for helping with the WSJ speech transcription pipeline. This work was funded by Samsung Institute of Advanced Technology, the University of Zurich and ETH Zurich. }

\ifCLASSOPTIONcaptionsoff
  \newpage
\fi

\bibliographystyle{IEEEtran}
\bibliography{main}

\end{document}